%% file: ecai-paper.tex
\documentclass{ecai}
\usepackage{graphicx}
\usepackage{latexsym}
\usepackage{amsmath}
\usepackage{algpseudocode}
\usepackage{algorithm}
\usepackage[title]{appendix}

\DeclareMathOperator*{\argmax}{argmax}

\begin{document}

\begin{frontmatter}

\title{Robust Assignment of Labels for Active Learning with Sparse and Noisy Annotations }

\author[A, B]{\fnms{Daniel}~\snm{Kałuża}\orcid{0000-0002-2544-5052}\thanks{Corresponding Author. Email: d.kaluza@mimuw.edu.pl}}
\author[A, B]{\fnms{Andrzej}~\snm{Janusz}\orcid{0000-0002-9763-1399}}
\author[A, B]{\fnms{Dominik}~\snm{Ślęzak}\orcid{0000-0003-2453-4974}}

\address[A]{Institute of Informatics, University of Warsaw, Poland}
\address[B]{QED Software, Poland}

\begin{abstract}
Supervised classification algorithms are used to solve a growing number of real-life problems around the globe. Their performance is strictly connected with the quality of labels used in training. Unfortunately, acquiring good-quality annotations for many tasks is infeasible or too expensive to be done in practice. 
To tackle this challenge, active learning algorithms are commonly employed to select only the most relevant data for labeling. However, this is possible only when the quality and quantity of labels acquired from experts are sufficient. Unfortunately, in many applications, a trade-off between annotating individual samples by multiple annotators to increase label quality vs. annotating new samples to increase the total number of labeled instances is necessary.
In this paper, we address the issue of faulty data annotations in the context of active learning. 
In particular, we propose two novel annotation unification algorithms that utilize unlabeled parts of the sample space. The proposed methods require little to no intersection between samples annotated by different experts. 
Our experiments on four public datasets indicate the robustness and superiority of the proposed methods in both, the estimation of the annotator's reliability, and the assignment of actual labels, against the state-of-the-art algorithms and the simple majority voting.
\end{abstract}

\end{frontmatter}

\section{Introduction}

Supervised learning algorithms are commonly used to create prediction models for a considered classification task. 
The quality of such models, for the majority of algorithms, strongly depends on the labeled dataset used during the model construction.
In real-life scenarios, we often start with no or only a few labeled samples, as the data annotation process is expensive and requires laborious human involvement. 
To optimize this process and cut costs, active learning algorithms are commonly employed~\cite{10.1145/3472291}. 

Active learning algorithms can be defined as follows - \textit{a set of algorithms that given a limited labeling budget try to obtain as best model as possible, assuming that they can iteratively query an oracle (usually human experts) to annotate chosen samples. 
}
In some cases, labels might be also obtained in an automated manner, e.g., by computer simulations. However, for many classification problems, e.g. security alert notifications~\cite{sod}, humans have to manually annotate chosen samples. For this reason, considerable domain knowledge and experience from human annotators is required, thus we usually refer to the annotators as~experts.

As humans are imperfect by nature, the acquired annotations might contain mistakes, influencing the quality of obtained models. The frequency of those mistakes usually depends on the difficulty of the task itself and the expertise of annotators. If those mistakes occur too often and the quality of acquired labels is insufficient, corrective measures have to be employed. In this field, there are two dominant approaches: annotations unification algorithms~\cite{raykar2010} (also known as consensus algorithms), and faulty labels identification and removal methods~\cite{KARIMI2020101759}. 

The first approach uses annotations from multiple human experts to assign a refined label to the sample, thus benefiting from the fact that some of the experts will assign it to the appropriate class. 
These methods require samples to be labeled by multiple experts and as the labeling budget is usually limited, enforce a trade-off between label quality and the quantity of labeled samples.
The second approach tries to identify and filter out mislabeled samples, which might also result in the removal of some correctly labeled instances. This is an undesirable side-effect, especially in low-budget annotation scenarios or in the case of imbalanced datasets. It leads to the oversimplification of the constructed model and losing important details about more complex data instances. 
Because of this risk, our work focuses on the first of the aforementioned groups, i.e. the label unification algorithms.

In this paper, we propose two algorithms based on the Expectation-Maximization technique (EM) and an intuitive idea to augment every expert using a machine learning model. 
A detailed description of our approach is available in Section \ref{algorithm}. 
Proposed algorithms infer labels for the whole data, and therefore do not have the major drawback of requiring many annotations per sample to achieve high-quality labels. 
We compare our methods with baseline reference, i.e. majority voting and commonly used EM-based algorithm in our experimental setup on four publicly available datasets. 
Our experiments are further described in Section \ref{experiments}. Since two of the datasets used in experiments are highly imbalanced, applying the most commonly used probability cut-off of 0.5 to assign labels leads to poor performance of models according to metrics adjusted for imbalanced classification, such as the balanced accuracy (BAC). To tackle this challenge, a novel cut-off computation method is proposed in Section \ref{prediction_algorithm}. The proposed method can be used even without prior knowledge about class distribution, which is suitable for typical active learning scenarios.

\section{Related Work}\label{related_work}

Reaching a consensus among labelers is one of the fundamental issues for active learning research~\cite{10.1145/3472291}. In this setting, the main objective is to iteratively select the most informative unlabeled samples and request their labels from an oracle, e.g., human annotators or other labeling sources. This approach has been successfully applied to various classification tasks such as text analysis and classification~\cite{shen-etal-2017-deep}, image classification~\cite{10.5555/3305381.3305504,Raczkowski2019}, and medical diagnosis~\cite{bressan2019breast,wu2021covid}. 
The most popular approach to the active selection of training instances is so-called, pool-based uncertainty sampling~\cite{10.1007/s10994-021-06003-9}. It assumes that there is an unlabelled pool of data available, from which an active learning algorithm can select the next batch of samples to be annotated by the oracle in the next labeling iteration. Data instances are chosen for labeling based on some estimation of the prediction uncertainty that can be computed using various approaches~\cite{BALD2017,Wu_2022_CVPR}.

Active learning has also been applied to many other types of prediction tasks, such as the multi-label classification, where each sample may belong to multiple classes simultaneously~\cite{10.5555/2540128.2540341}. In this case, the annotation process becomes even more complex, as multiple labels need to be assigned to each sample. Approaches that have been proposed to address this issue include models investigating correlations between label occurrences or methods that select samples based on the uncertainty of the whole label set. These methods have been shown to be effective in reducing the labeling cost and improving the performance of multi-label classification models~\cite{10.1145/1291233.1291245}. However, active learning has also been successful for regression problems~\cite{DBLP:conf/bigdataconf/KaluzaJS22} and many other ML tasks. A comprehensive survey of active learning applications and sample selection techniques can be found in~\cite {10.1145/3472291}.

In practice, annotations provided by human labelers quite often contain errors or inconsistencies which can negatively impact the performance of active learning algorithms. A number of research papers have addressed this issue by proposing annotation aggregation methods that can improve the quality of labels. For example, there are methods that combine multiple annotations using majority voting~\cite{10.1145/1401890.1401965} or EM-based algorithms~\cite{raykar2010}. Some of the recent approaches include learning-based methods that incorporate information about annotator expertise to improve annotation quality~\cite{10.5555/3504035.3504216}. An example of such an approach is the multi-label consensus maximization for ranking (MLCM-r) algorithm proposed by~\cite{6729628}.
Another example is the Dawid-Skene model~\cite{Skene18}. It assumes that annotators have different error rates for different decision classes and models the probability of a correct label for each sample, given the annotations provided by multiple annotators. Additionally, it uses the EM algorithm to estimate the true labels of the samples and the reliability of each annotator. Several studies have demonstrated the effectiveness of these approaches in reducing the impact of noisy annotations on the performance of active learning algorithms~\cite{Shahana-crowdsourcing} and in scenarios when the federated learning techniques were applied~\cite{Zeng-federated-learning-2022}.

\section{Annotations Unification Algorithms} \label{algorithm}

In this section, we delve into the details of proposed algorithms denoted as inferred consensus and simulated consensus algorithms.
We consider simulated consensus as a more stable and refined version of the inferred consensus algorithm, however, we present both to comprehensively describe the intuition behind them.
Both proposed algorithms have been developed to overcome a major drawback of consensus algorithms, i.e. degradation of performance if many samples are not labeled by multiple experts. We have developed them as extensions of the EM algorithm, as it is the most well-known consensus algorithm, tested in many production implementations. Actually, proposed extensions are independent of the EM itself and can be viewed as meta-techniques. We are convinced they might also be used with other annotation unification algorithms and lead to a refined performance in many circumstances. However, as our experiments cover only the case when they are used together with EM, we will describe them in that context in the rest of this section.

\subsection{Expectation-maximization}
The application of the EM algorithm to the task of estimation of the labels based on multiple noisy annotations has been originally proposed by Raykar et al. \cite{raykar2010}. It was shown to be a robust solution when labels are abundant. Here we briefly paraphrase the theory for binary classification, but it can also be easily used for multi-label scenarios which can be modeled as multiple binary classifications or extended to multi-class problems as shown in the original paper.

Let us denote a true label of the sample $i$ as $y_i$, a label assigned to this sample by expert $j$ as $y_i^j$, a representation of this sample as $x_i$, the number of all samples as $N$, and the number of all experts as $R$. As this work focuses on sparse annotations, we will denote indices of samples annotated by expert $j$ as $S^j \subseteq \{1, \ldots, N\}$, and symmetrically the set of experts that have labeled sample $i$ as $E_i \subseteq \{1, \ldots, R\}$.

This probabilistic algorithm makes the following simplifications:
\begin{itemize}
    \item Each expert $j$ is modeled by two latent variables measuring expertise for the given class, namely specificity (true negative rate) $\beta^j$ and sensitivity (true positive rate) $\alpha^j$.
    \item Probability that an expert assigns a specific class to the sample depends only on the true hidden label of this sample and latent variables of this expert. In other words, they do not depend on the representation of this sample given the true label. I.e.: 
    \begin{equation*}
        P(y_i^j = 1 | x_i, y_i) = P(y_i^j = 1 | y_i).
    \end{equation*}
    \item Each expert annotates samples independently from other annotators, therefore assigned classes are independent given the true labels.
    \begin{equation*}
        P(y_i^j = 1 | y_i, y_i^k) = P(y_i^j = 1 | y_i) \quad \text{if} \, j \neq k.
    \end{equation*}
\end{itemize}

The EM algorithm starts by initializing the first estimated probability of true labels with majority voting and then iteratively repeats E and M steps until convergence to stable parameters and probability estimation of true labels.

\subsubsection{E-step}

We will denote the set of all learned parameters of the algorithm as $\Theta$, containing $\alpha, \beta$, and the parameters of the machine learning model if one is used for posterior probability estimation.
Then, based on the independence of the annotators given a true label and Bayes' theorem, the probability of a positive class $\mu_i = P(y_i = 1 | y_i^1, ... y_i^R, \Theta, x_i )$\\ can be written as:

\begin{align}
\mu_i =  \frac{P( y_i^1, ... y_i^R | y_i = 1, \Theta) \cdot P(y_i = 1 | \Theta, x_i) }{P( y_i^1, ... y_i^R | \Theta, x_i)} \\ 
\propto P( y_i^1, ... y_i^R | y_i = 1, \Theta) \cdot P(y_i = 1 | \Theta, x_i).
\end{align}

Where $P(y_i = 1 | \Theta, x_i)$ is posterior probability and can be modeled with a machine learning model, we will denote it as $p_i$. $P( y_i^1, ... y_i^R | \Theta, x_i)$ does not depend on the label, therefore it is of no interest to us and can be handled by normalization of scores to a proper probability distribution. If we define $a_i = P( y_i^1, ... y_i^R | y_i = 1, \alpha)$ and $b_i = P( y_i^1, ... y_i^R | y_i = 0, \beta)$, we can rewrite equation for $\mu_i$ to: 

\begin{eqnarray}
    \mu_i & = & \frac{a_i p_i}{a_i p_i + b_i (1 - p_i)}, \\
a_i & = & \prod_{j \in E_i} 
    [\alpha^j]^{y_i^j} 
    [1 - \alpha^j]^{(1-y_i^j)}, \\
b_i & = & \prod_{j \in E_i} 
    [\beta^j]^{(1-y_i^j)} 
    [1-\beta_i^j]^{y_i^j}.
\end{eqnarray}

The last set of equations can be used to efficiently compute the expected probability for the positive class.

\subsubsection{M-step}

The maximization step is used to update parameters $\Theta$ of the algorithm.
The equations resulting from computing the gradient of log-likelihood of estimated labels over the parameters $\alpha, \beta$ look as follows:

\begin{eqnarray}
    \alpha^j  & = &  \frac{\sum_{i \in S^j} \mu_i y_i^j} {\sum_{i \in S^j} \mu_i} \\
    \beta^j  & = &  \frac{\sum_{i \in S^j} (1 - \mu_i) (1 - y_i^j)} {\sum_{i \in S^j} (1 - \mu_i)}.
\end{eqnarray}

An update of the parameters of a machine learning model used for the posterior probability prediction can be done using the regular gradient descent method.

\subsection{Inferred consensus} \label{inferred_consensus}
As the performance of the EM algorithm degrades with smaller numbers of annotations for each sample, the  main idea of inferred consensus algorithm is to propagate the annotations to unlabelled samples, using the knowledge from the samples that an expert has labeled. The intuition behind this idea is expressed by the following question: "What label do we expect annotator $j$ would have given sample $i$, which hasn't been annotated by him?"
To be able to answer this question and infer the predictions, for every expert a machine learning model is trained on annotations given by this expert. 

More formally, for expert $j$ we create a model $f^j$ trained on samples $<x_i,y_i^j>_{i \in S_i}$. Then, this model is used to infer predictions for the whole dataset obtaining new annotations, $y'^j_i = f^j(x_i)$ for $i \in \{1, ..., N \}$ and every expert $j \in \{1, ..., R\}$. As the majority of machine learning models return not only a label but also a probability distribution of classes, we utilize the returned distribution as soft annotations, e.g., an artificial expert says that from its perspective there is a 10\% chance that the object has the positive class and 90\% chance it belongs to the negative class. 
Finally, the EM algorithm can be run on inferred annotations $y'$ for all of the samples,  potentially leading to a better estimation of the hidden true labels, as we have a full inferred annotation set of size $R$ for every sample.

The algorithm can be presented as the following set of steps:

\begin{enumerate}
    \item Train machine learning model $f^j$ for each expert using  $<x_i,y_i^j>_{i \in S_i}$.
    \item Infer predictions $y'^j_i = f^j(x_i)$ for $i \in \{1, ..., N \}$
    \item Call EM algorithm using $y'$ instead of original annotations.
\end{enumerate}

This algorithm can be viewed as the creation of a new labeling task, that was annotated by artificial experts derived from the original annotators. The advantage of this task is that it is fully labeled by each annotator, therefore it is more suitable for the EM algorithm, and the downside is that artificially created annotators usually have worse quality than original experts, as they are trained only on the small subset of samples, and dependant on the used machine learning model. Moreover, since we associate real experts with models trained on samples annotated by them, we obtain unreliable estimations of experts' reliability, which changes during the annotation process, as the model usually gets better with the increasing number of samples annotated by the expert.

\subsection{Simulated consensus} \label{simulated_consensus}

\begin{figure*}
\centerline{\includegraphics[height=2.1in]{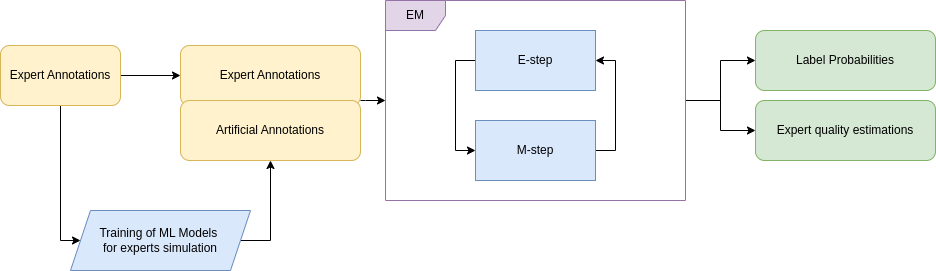}}
\caption{Visualization of Simulated consensus algorithm steps. Algorithm outputs are denoted with green, annotations data yellow and algorithm steps are shown as blue.} \label{simulated_consensus_fig}
\end{figure*}

To fix downsides of the inferred consensus algorithm we have prepared a more mature and refined version called simulated consensus.
The schematic illustration of the algorithm can be seen on Figure \ref{simulated_consensus_fig}.
Once again we start  by training a machine learning  model for each expert, but now we infer predictions only on samples that has not been annotated by this expert, i.e. were not used in training of this model. 
Then, we use the predictions (in form of probability distributions) as annotations from a new expert fully separate from the original one. In this way we obtain $2R$ annotators, when first $R$ of them are human experts, and second $R$ are simulated. Finally, the EM algorithm is used on the combined, partially soft, annotations set.   

The algorithm works as follows:
\begin{enumerate} 
    \item Train machine learning model $f^j$ for each expert using  $<x_i,y_i^j>_{i \in S_i}$.
    \item Infer predictions $y'^j_i = f^j(x_i)$ for $i \notin S_i$
    \item Create new annotations $\hat{y}$ as concatenation of $y^j|_{j \in \{1, ..., R\}}$ and $y'$ 
    \item Call EM algorithm using $\hat{y}$ instead of original annotations.
\end{enumerate}

This algorithm also leads to performing consensus on a set of $R$ annotations for each sample, therefore tackling major drawback of original EM in case of sparse annotations. Moreover, it has several advantages over inferred consensus algorithm from the theoretical point of view. First of all, we are using original annotations of the experts and, as they are fully separated from the artificially created annotators, we reliably evaluate their quality. We also believe that the quality of the experts might be better evaluated as there is always quorum of $R$ annotators participating in voting for each sample. Besides that, the algorithm is not so prone to errors because of poor quality of created machine learning models, because their quality is also separately evaluated in EM (we also expect this to be a decent evaluation as none of the artificial experts make predictions on its training samples) and if they achieve poor performance their influence in the voting diminishes. 

Intuitively this algorithm also can be viewed as a new labelling task in the following mind experiment. Lets imagine that we have a joint set of original experts annotations and another group of slightly worse artificial annotators. In real world, there might exist a person who would return the same annotations as our artificial annotator. Therefore, those should be perfectly fine annotations from the perspective of annotations unification algorithm, and if they are of poor quality algorithm should evaluate them as such and be only slightly guided by them.

\section{Prediction for imbalanced data}\label{prediction_algorithm}

Expectation-Maximization algorithm results in estimation of probability distribution of class labels for each sample, but many models cannot be trained using such soft labels. Usually, we humans also expect a definitive answer whether sample should be considered as belonging to particular class or not. If considered machine learning task corresponds to balanced machine learning distribution a standard $0.5$ cut-off for binary classification or $\frac{1}{K}$, where $K$ is the number of classes, is a sound solution. However, if we are dealing with imbalanced classification and try to optimize metric, which attaches the same importance to recognition of each class like balanced accuracy, it is not a good threshold. 

In some active learning scenarios with noisy annotations even an approximated class distribution might not be known a priori, e.g. cybersecurity attacks detection. In such cases, threshold tuning is infeasible, as we do not have a reliable validation set to evaluate the threshold efficiency. Therefore, a reliable method is needed nor requiring prior knowledge nor true labels. This is why we propose the following method of class distribution approximation using available model.

\begin{enumerate}
    \item Compute a probability distribution from the perspective of the model for all samples available during training, lets call a distribution for sample $i$ a $\tilde{y}_i$ and probability of class $c$ for this sample $\tilde{y}_{i,c}$.
    \item Compute mean probability for each class across all of the samples. This mean probability will be a threshold adjustment lets call it $t_c$ for class $c$. Formally:
    \begin{equation*}
        t_c = \frac{1}{N} \sum_{i=1}^N \tilde{y}_{i,c}.
    \end{equation*}
    \item For multi label classification we assign class $c$ to the sample $i$ if $\tilde{y}_{i,c} \geq t_c$.
    \item For single label classification we choose class $c'$ from the set of all classes $C$ in the following way: 
    \begin{equation*}
            c' = \argmax_{c \in C} \tilde{y}_{i,c} - t_c.
    \end{equation*}
\end{enumerate}

This method allows us to determine the cut-off without any prior knowledge about the problem. It can be used for EM algorithms, then considered model is the EM itself, and predictions are the resulting probability distribution. Moreover, it can also be used to choose a threshold for machine learning models trained on top of that. In such case, the $t_c$ is computed on all pool data available during training, over the predictions of the model we wish to threshold. Thanks to this fact, no additional computation is needed in production and the adjustment is independent from the data set we are making the prediction on.

When we use this procedure on an unbiased model, it leads to an unbiased estimator of true class distribution. Moreover, for balanced datasets, it converges with the increasing number of training samples to the regular $\frac{1}{K}$ threshold.

\section{Experiments}\label{experiments}

\subsection{Experimental setup}
To properly evaluate proposed algorithms, we have created an experimental setup similar to a real-life active learning scenario.
As data labeling by human experts only for the purpose of experiments is too expensive and in real-life scenarios with annotators you usually do not have access to the hidden true labels, we have prepared a randomized procedure for creating annotations based on true labels of known public datasets.
The procedure generates a set of binary annotations for a specified number of experts, which is a parameter of the method, in the following~way:
\begin{enumerate}
    \item Number of labeled samples differs for each expert. We model the probability that expert $j$ annotates a sample, denoted as $r^j$, as a Beta distribution with parameters $\tilde{\alpha} = 1$,  $\tilde{\beta} = 20$, therefore average probability is equal to $\frac{1}{21} \approx 0.048$. Intuitively, we can think of it that experts will on average label one per every 21 samples.
    \item The fact that an expert $j$ has annotated a sample $i$ is decided by drawing from Bernoulli distribution with a success ratio equal~$r^j$. 
    \item The hidden true positive and true negative rates of each expert $j$, denoted as $\hat{\alpha}^j$ and  $\hat{\beta}^j$, are drawn from the Beta distribution with parameters $\tilde{\alpha} = 4$,  $\tilde{\beta} = 1$, that have the expected value equal~$0.8$.
    \item Classes assigned by the expert $j$ to annotated sample $i$ are drawn from Bernoulli distribution with the probability of success $\hat{\alpha}^j$ if the true label of this sample is positive or is equal to $1 - v$, where $v$ is drawn from Bernoulli with the probability of success equal to $\hat{\beta}^j$  when the true label of the sample $i$ is negative.
\end{enumerate}

We set the number of experts to 15 for our experiments because the randomization of samples annotated by each expert might lead to experts labeling only a few samples (which is consistent with a real-life scenario when somebody leaves a company after a few days of annotation).
Using the above procedure, we obtained annotations assigned by diverse artificial experts. Thanks to the fact that it is based on public datasets, we had true labels for all of the samples and the hidden quality of each expert to properly evaluate tested algorithms. 
As the proposed annotation generation procedure assumes the binary classification task and works in an independent manner for each class, we used one-hot encoded labels of every problem as in a multi-label setting.  
The evaluation for each dataset was performed five times to obtain the statistical significance of the experiments, each time with a fixed seed creating a different set of expert annotations. The evaluation procedure was as follows:

\begin{enumerate}
    \item Create expert annotations for a given random seed.
    \item Use each consensus method to generate label probabilities and experts' quality estimations.
    \item Generate labels using all tested cut-off techniques.
    \item Train a machine learning model on the obtained labels. 
    \item Make predictions with the obtained model on the separate test set. Use the cut-off techniques again to assign labels to test cases.
    \item Compute evaluation metrics, both on the consensus results and the model predictions. 
\end{enumerate}

We have included a quality assessment of resulting machine learning models, trained on the obtained labels, as this is usually the ultimate result of an active learning system. If a sophisticated consensus method led to a better estimation of labels but would not lead to a better machine learning model, there would be no advantage in using this method in a production environment.
To reduce the computational complexity, a machine learning model used for posterior probability distribution prediction inside EM-based algorithms (that has to be retrained with every iteration) was a dummy model predicting always the class prior probability estimated on the training set regardless of the passed sample. Nevertheless, a regular  machine learning model chosen for each task was trained on top of computed labels for the evaluation.  
 
\subsection{Evaluation metrics}

The evaluation metrics used in our experiments can be divided into three groups. 

In the first group, there are metrics computed on the probabilities from the annotations unification algorithms. All of these metrics are computed on the set of samples that were annotated by at least one expert in the experiment. 
We considered the following measures: area under the receiver operating characteristic curve (AUC) with the macro average on the probabilities returned by the algorithms, and balanced accuracy (BAC) on the labels generated by each of the cut-off methods.

The second group contains evaluations of the estimated quality of each expert by the compared algorithms. Metrics used in the comparison: the mean absolute error of the true positive rate estimation (MAE), Pearson correlation, and the Spearman rank correlation between the estimated true positive rates and the hidden true positive rates.

The third group contains evaluations of the machine learning models trained on estimated labels. A separate model is trained for each consensus method and each cut-off technique. For each model, a BAC score on the test set is reported. As BAC requires labels and the models return probability scores, the same cut-off method as was used to generate training labels is applied.

\subsection{Datasets}

We have used four datasets for the purpose of evaluation.

\begin{itemize}
    \item MNIST \cite{mnist} - A dataset of handwritten digits, one of the most widely used benchmark datasets in machine learning research.
    \item firefighters \cite{aaia2015} - A dataset with measurements from wearable inertial sensors placed on fire-fighters during various fire and rescue-related activities from the \emph{AAIA'15 Data Mining Competition: Tagging Firefighter Activities at a Fire Scene} organized at the KnowledgePit.ai platform. 
    \item cybersec \cite{sod} - A dataset describing cybersecurity network logs with a prediction task to identify events that should be notified as suspicious. This dataset was originally published in a competition \emph{IEEE BigData 2019 Cup: Suspicious Network Event Recognition} on the KnowledgePit.ai platform. 
    \item credit-fraud \cite{credit_fraud} - A public dataset of transactions made by European credit card holders, fully anonymized via PCA transformation. The dataset is publicly available both in the OpenML repository and on the Kaggle competition platform. The prediction task is to detect fraudulent transactions. 
\end{itemize}

Those datasets were chosen to diversify both, domains and class distributions used to evaluate our methods. MNIST is a balanced dataset with ten classes, firefighters data have five classes with slightly imbalanced distribution, cybersec is a binary classification task and has imbalanced distribution with less than $6\%$ of positive samples, and credit-fraud is a binary and highly imbalanced dataset with less than $1\%$ fraud examples.
Moreover, all of these datasets required human annotations at some point to create the labels for the corresponding tasks. We cannot be sure whether there are errors in the labels, but such investigation remains outside of the scope of this study. 
For both MNIST and credit-fraud, test sets for evaluation were created by a stratified split with $40\%$ of all available samples, whereas for cybersec and firefighters, splits from the corresponding data science competitions were used. Moreover, for the cybersec and firefighters datasets, the same preprocessing as described in the referenced competition papers was performed. Additionally, each dataset was min-max scaled.

Model architectures with hyper-parameters used for evaluation are shown in Table \ref{model_params}. For MNIST and firefighters, a logistic regression model with default parameters was used. For cybersec and credit-fraud, the XGBoost classifier was used. Since those are highly imbalanced datasets, an appropriate scaling parameter was used with a value equal to the ratio of negative and positive samples was used for training the models.

\subsection{Consensus methods and cut-off threshold}

In our experiments we have evaluated the following consensus methods:

\begin{itemize}
    \item Simulated consensus - a refined version of the proposed algorithm generating additional annotations for each sample with machine learning models described in Section \ref{simulated_consensus}.
    \item Inferred consensus -  the first revision of the proposed algorithm, substituting expert annotations with machine learning models described in detail in Section \ref{inferred_consensus}.
    \item EM - the original expectation-maximization algorithm.
    \item Majority voting - the regular majority voting algorithm with a slight modification to make it more comparable with other methods. The modification is as follows - it returns a distribution of votes for individual classes instead of just indicating the class with the highest number of votes.   
\end{itemize}

In both, inferred consensus and simulated consensus, models representing experts had exactly the same architecture and hyperparameters as the final model used in the evaluation. The parameter values are given in Table \ref{model_params}.

\begin{table}
\begin{center}
{\caption{Machine learning models and their relevant hyperparameters used for each of the machine learning tasks. Default hyperparameters have been omitted, XGBoost library in version 1.6.2 and scikit-learn 0.24.2 were used to train the models. }\label{model_params}}
\begin{tabular}{|c|c|c|}
\hline

Dataset & Model & Hyperparameters 
\\
\hline
 MNIST & Logistic Regression & max\_iter=500, n\_jobs=10\\
 \hline
 firefighters & Logistic Regression & max\_iter=500, n\_jobs=10 \\
 \hline

 cybersec & XGBClassifier & neg\_pos\_ratio=$\frac{\#neg}{\#pos}$,  \\ 
 & &  n\_estimators=300, max\_depth=3, \\
 & &   learning\_rate=0.05, n\_jobs=10 \\
 \hline
 credit-fraud & XGBClassifier & neg\_pos\_ratio=$\frac{\#neg}{\#pos}$,  \\ 
 & &  n\_estimators=300, max\_depth=3, \\
 & &   learning\_rate=0.05, n\_jobs=10 \\
 \hline
\end{tabular}
\end{center}
\end{table}

The following cut-off thresholding techniques were used:
\begin{itemize}
    \item Default - Default $0.5$ threshold used in the majority of machine learning frameworks.
    \item GT-prior - A threshold computed using true labels from the training pool. This threshold represents the ratio of samples having a particular class to all of the samples in the pool.
    \item Model-posterior - The proposed thresholding technique that uses the probability distribution predicted for the whole available training data pool, as described in Section \ref{prediction_algorithm}. Keep in mind that for each model, the prediction was done over all available samples from the pool, not only those which were annotated by experts. 
\end{itemize}

Those cut-off thresholds were used to generate labels in the same way as described in Section \ref{prediction_algorithm}. For the purpose of multi-label model training, a probability distribution was compared with the corresponding threshold to determine whether a class should be assigned to the sample. For the BAC estimation, a difference between the maximal predicted probability and the threshold value was~used. 

\section{Results}\label{results}

\begin{table*}
    
\begin{center}
{\caption{Results of annotation quality metrics, each dataset has a separate subsection. Each row features the results of one annotation unification method for the corresponding dataset. The first column named AUC denotes the area under the ROC curve computed between obtained probabilities and true labels for annotated samples. The rest of the columns denote the balanced accuracy between labels obtained with the thresholding method indicated in the column name and true labels for annotated samples. Standard deviations across the experiments are shown in brackets next to each value. Bold values indicate the largest value in the AUC column and across all BAC columns for each of the datasets. }\label{annotations_quality}}
\begin{tabular}{|c|c|c|c| c |}
\hline

Method & AUC & BAC-default & BAC-GT-prior & BAC-model-posterior \\

\hline
 \multicolumn{5}{|c|}{MNIST} \\
\hline

Simulated consensus & $  \mathbf{ 0.988 (\pm  0.002)} $ & $ \mathbf{ 0.911 (\pm  0.010)} $ & $  0.908 (\pm  0.010) $ & $  0.907 (\pm  0.011) $ \\
Inferred consensus & $  0.978 (\pm  0.001) $ & $  0.870 (\pm  0.004) $ & $  0.868 (\pm  0.004) $ & $  0.867 (\pm  0.005) $ \\
EM & $  0.882 (\pm  0.018) $ & $  0.588 (\pm  0.033) $ & $  0.589 (\pm  0.034) $ & $  0.590 (\pm  0.034) $ \\
Majority Voting & $  0.801 (\pm  0.008) $ & $  0.405 (\pm  0.030) $ & $  0.419 (\pm  0.020) $ & $  0.459 (\pm  0.024) $ \\
\hline
 \multicolumn{5}{|c|}{firefighters} \\
\hline

Simulated consensus & $  0.979 (\pm  0.009) $ & $  0.872 (\pm  0.046) $ & $  0.874 (\pm  0.042) $ & $  \mathbf{ 0.875 (\pm  0.042) } $ \\
Inferred consensus & $ \mathbf{ 0.985 (\pm  0.003)} $ & $  0.845 (\pm  0.051) $ & $  0.840 (\pm  0.047) $ & $  0.842 (\pm  0.047) $ \\
EM & $  0.875 (\pm  0.027) $ & $  0.647 (\pm  0.053) $ & $  0.681 (\pm  0.040) $ & $  0.687 (\pm  0.036) $ \\
Majority Voting & $  0.798 (\pm  0.016) $ & $  0.581 (\pm  0.034) $ & $  0.569 (\pm  0.041) $ & $  0.573 (\pm  0.037) $ \\
\hline
 \multicolumn{5}{|c|}{cybersec} \\
\hline

Simulated consensus & $ \mathbf{ 0.909 (\pm  0.022)} $ & $  0.635 (\pm  0.054) $ & $  \mathbf{ 0.887 (\pm  0.020)} $ & $  0.873 (\pm  0.019) $ \\
Inferred consensus & $  0.784 (\pm  0.131) $ & $  0.500 (\pm  0.001) $ & $  0.556 (\pm  0.073) $ & $  0.729 (\pm  0.022) $ \\
EM & $  0.876 (\pm  0.021) $ & $  0.821 (\pm  0.027) $ & $  0.515 (\pm  0.029) $ & $  0.827 (\pm  0.022) $ \\
Majority Voting & $  0.797 (\pm  0.023) $ & $  0.805 (\pm  0.025) $ & $  0.789 (\pm  0.041) $ & $  0.789 (\pm  0.041) $ \\
\hline
 \multicolumn{5}{|c|}{credit-fraud} \\
\hline

Simulated consensus & $ \mathbf{ 0.869 (\pm  0.045)} $ & $  0.538 (\pm  0.035) $ & $  0.683 (\pm  0.178) $ & $  \mathbf{ 0.838 (\pm  0.057)} $ \\
Inferred consensus & $  0.747 (\pm  0.151) $ & $  0.500 (\pm  0.000) $ & $  0.628 (\pm  0.159) $ & $  0.769 (\pm  0.145) $ \\
EM & $  0.801 (\pm  0.061) $ & $  0.616 (\pm  0.059) $ & $  0.500 (\pm  0.000) $ & $  0.781 (\pm  0.052) $ \\
Majority Voting & $  0.807 (\pm  0.045) $ & $  0.810 (\pm  0.041) $ & $  0.804 (\pm  0.051) $ & $  0.804 (\pm  0.051) $ \\
\hline

\end{tabular}
\end{center}
\end{table*}

\begin{table*}
    
\begin{center}
{\caption{ Results of experts' quality estimation metrics. Each dataset has a separate subsection. Each row features results for one annotation unification method for the corresponding dataset. The first column, named MAE, denotes the mean absolute error across estimations of true positive rates for experts. Pearson and Spearman indicate values of the corresponding correlation coefficients between the estimated true positive rates and the ground truths assigned during the experiment setup. Standard deviations across the experiments are shown in brackets next to each value. Bold values indicate the smallest MAE or the largest correlation for each of the datasets. }\label{expert_estimation_quality}}
\begin{tabular}{|c|c|c|c|}
\hline

Method & MAE & Pearson & Spearman \\

\hline
 \multicolumn{4}{|c|}{MNIST} \\
\hline

Simulated consensus & $  \mathbf{ 0.045 (\pm  0.009)} $ & $  \mathbf{ 0.902 (\pm  0.044)} $ & $  \mathbf{ 0.894 (\pm  0.051)} $ \\
Inferred consensus & $  0.175 (\pm  0.009) $ & $  0.775 (\pm  0.063) $ & $  0.763 (\pm  0.065) $ \\
EM & $  0.090 (\pm  0.020) $ & $  0.757 (\pm  0.077) $ & $  0.671 (\pm  0.114) $ \\
Majority Voting &   NA  & NA & NA \\
\hline
 \multicolumn{4}{|c|}{firefighters} \\
\hline

Simulated consensus & $  \mathbf{ 0.083 (\pm  0.013)} $ & $  \mathbf{ 0.689 (\pm  0.109)} $ & $  \mathbf{ 0.700 (\pm  0.096)} $ \\
Inferred consensus & $  0.179 (\pm  0.012) $ & $  0.608 (\pm  0.088) $ & $  0.677 (\pm  0.056) $ \\
EM & $  0.122 (\pm  0.028) $ & $  0.567 (\pm  0.149) $ & $  0.566 (\pm  0.190) $ \\
Majority Voting &   NA  & NA & NA \\

\hline
 \multicolumn{4}{|c|}{cybersec} \\
\hline

Simulated consensus & $  \mathbf{ 0.065 (\pm  0.015)} $ & $  \mathbf{ 0.756 (\pm  0.129)} $ & $  \mathbf{ 0.713 (\pm  0.220)} $ \\
Inferred consensus & $  0.275 (\pm  0.073) $ & $  0.358 (\pm  0.230) $ & $  0.408 (\pm  0.280) $ \\
EM & $  0.101 (\pm  0.032) $ & $  0.689 (\pm  0.226) $ & $  0.634 (\pm  0.175) $ \\
Majority Voting &   NA  & NA & NA \\

\hline
 \multicolumn{4}{|c|}{credit-fraud} \\
\hline

Simulated consensus & $  \mathbf{ 0.126 (\pm  0.053)} $ & $  \mathbf{ 0.456 (\pm  0.261)} $ & $  \mathbf{ 0.448 (\pm  0.199)} $ \\
Inferred consensus & $  0.268 (\pm  0.053) $ & $  0.164 (\pm  0.178) $ & $  0.221 (\pm  0.147) $ \\
EM & $  0.250 (\pm  0.025) $ & $  0.211 (\pm  0.162) $ & $  0.253 (\pm  0.122) $ \\
Majority Voting &   NA  & NA & NA \\

\hline

\end{tabular}
\end{center}
\end{table*}

\subsection{Annotations quality}

A summary of annotation quality results can be found in Table \ref{annotations_quality}. The simulated consensus algorithm has obtained significantly better results than all other methods on all datasets but firefighters in both ROC AUC and BAC metrics. On the firefighters dataset, the inferred consensus obtained slightly better ROC AUC than the simulated consensus, which turned out to be the second for this metric. Moreover, we computed the one-sided Wilcoxon signed rank test to check the statistical significance of these results. Scores obtained by the simulated consensus turned out to be significantly greater than the scores of the EM algorithm for all of the datasets in terms of both AUC and BAC-model-posterior with a p-value of $0.03125$, which we consider a good result taking into account the limited expressiveness of Wilcoxon test. These results show the robustness and superiority of the proposed annotation unification algorithm.

Noteworthy are also the results of the BAC-model-posterior cut-off, which obtained comparable performance for the balanced datasets and  better results than the default threshold for most of the consensus methods on imbalanced dataset combinations. For some imbalanced cases (cybersec and credit-fraud for the inferred consensus and the EM method), it led to good quality labels even when all other cut-off strategies failed. Tested using the one-sided Wilcoxon singed rank test against the default cut-off method, it obtained p-values: $0.026$, and $0.001$ for the cybersec and credit-fraud datasets, respectively. It suggests that this technique, which does not require a priori knowledge about label distribution, is the safest choice for new active learning scenarios.   

\subsection{Expert's reliability estimation}

Results of experts' true positive rate estimations are shown in Table \ref{expert_estimation_quality}. As suspected, the proposed inferred consensus method leads to distortion of expert reliability estimation. Therefore, it obtains larger mean absolute errors than the regular EM algorithm. Interestingly, the inferred consensus still results in greater correlations for the MNIST and firefighters datasets, which might be caused by better estimation of actual labels. 

Nevertheless, the refined version of our algorithm, i.e. simulated consensus, achieves highly superior scores in all three metrics and for all of the datasets. The p-values of the one-sided Wilcoxon rank test were: $0.03125$, $0.06250$, $0.31250$, and $0.03125$ for MNIST, firefighters, cybersec, and credit-fraud, respectively. The same p-values were obtained for both correlation metrics. Similar results were obtained for MAE: $0.03125$, $0.03125$, $0.09375$, and $0.03125$ for the corresponding datasets. Therefore, leading to statistically significant differences in two datasets for correlations and for three datasets for MAE. Noteworthy is the fact that due to the relatively small number of experiment repetitions, the expressiveness of the statistical test was severely limited. However, it still shows the potential of our method considering the fact that the MAE metric on MNIST and credit-fraud datasets was two times smaller on average than for other~methods.  

\subsection{Quality of trained models}

Results of model-related metrics can be found in the supplementary materials Appendix A. Our methods led to better machine learning models on the MNIST and firefighters datasets. The simulated consensus model achieved BAC of $ 0.878 (\pm  0.003)$ with model-posterior cut-offs technique on the MNIST dataset and the inferred consensus model achieved BAC of $ 0.791 (\pm  0.012) $, also model-posterior cut-offs, on the firefighters dataset. This finding is consistent with the label quality results. Surprisingly, on both imbalanced datasets classical majority voting with the default 0.5 threshold achieved better performance than any other model, i.e., $0.773(\pm 0.015)$, and $0.758(\pm 0.019)$ for the cybersec and credit-fraud datasets, respectively. This result is interesting, as other methods have obtained distinctively better label quality estimations on those datasets. The 0.5 threshold on model predictions looks sound from our perspective, as those models were trained with scaled weights for each class to balance the training data, however, we do not have a good explanation for why this threshold is also good for assigning labels for the majority voting algorithm. Therefore, as there is no clear correlation between labels and the resulting model's quality for the imbalanced datasets, this remains a topic for future research.

\section{Conclusions}

In this paper, we have addressed the issue of faulty data annotations in the context of active learning for classification. We proposed two novel annotation unification algorithms based on Expectation-Maximization (EM) and machine learning models, which require little to no intersection between samples annotated by different experts. Our experiments on four public datasets showed that the proposed methods outperform the state-of-the-art algorithms and simple majority voting, both in terms of the estimation of annotator reliability and the assignment of actual labels.
We also proposed a novel cut-off method to tackle the challenge of imbalanced datasets, which can be used even without prior knowledge about class distribution. This approach can be useful in many active learning scenarios where the distribution of classes is unknown or changes over time.

In conclusion, our proposed methods offer an effective solution to the issue of faulty data annotations. By utilizing unlabeled parts of the sample space and incorporating machine learning models, we can improve the quality of labeled datasets and ultimately enhance the performance of supervised classification algorithms. We hope that our work will contribute to further advancements in this field and encourage more research on consensus algorithms for data labeling.

Moreover, our research opens new, as far as we know, yet unexplored topics. Namely, one may ask why for some datasets labels quality does not clearly correlate with the quality of trained machine learning models. 
As this has a strong influence on all actively annotated machine learning tasks, it requires additional investigation in the future. Of course, this observation might be a result of a relatively small number of experiments, therefore to properly confirm the findings of this paper additional experiment repetitions and validation on new datasets are needed.

\ack This research was co-funded by Smart Growth Operational Programme 2014-2020, financed by European Regional Development Fund, in frame of project POIR.01.01.01-00-0213/19, operated by National Centre for Research and Development in Poland. 

\bibliography{ecai}

\appendix
\input{suplementary-materials}

\end{document}

%% file: suplementary-materials.tex
\onecolumn

\begin{appendices}
\section{Model quality results}\label{model_quality}

\begin{table*}[h]
    
\begin{center}
{\caption{Results of model quality metrics, each dataset has a separate subsection. Each row features results of models trained on labels from one annotation unification method for appropriate dataset. All of the columns denote balanced accuracy between labels obtained with thresholding method indicated in the column name applied to model predictions and true labels for test samples. Standard deviations across the experiments are show in brackets next to each value. Bold values indicate the largest value across all BAC columns for each of the datasets. }}
\begin{tabular}{|c|c|c|c|}
\hline

Method &  BAC-default & BAC-GT-prior & BAC-model-posterior \\

\hline
 \multicolumn{4}{|c|}{MNIST} \\
\hline
Simulated consensus & $  0.877 (\pm  0.005) $ & $  0.876 (\pm  0.004) $ & $  \mathbf{ 0.878 (\pm  0.003)} $ \\
Inferred consensus & $  0.863 (\pm  0.006) $ & $  0.860 (\pm  0.003) $ & $  0.865 (\pm  0.003) $ \\
EM & $  0.787 (\pm  0.040) $ & $  0.802 (\pm  0.020) $ & $  0.818 (\pm  0.015) $ \\
Majority Voting & $  0.798 (\pm  0.017) $ & $  0.800 (\pm  0.016) $ & $  0.817 (\pm  0.016) $ \\
\hline
 \multicolumn{4}{|c|}{firefighters} \\
\hline
Simulated consensus & $  0.738 (\pm  0.027) $ & $  0.750 (\pm  0.013) $ & $  0.750 (\pm  0.017) $ \\
Inferred consensus & $  0.733 (\pm  0.035) $ & $  0.763 (\pm  0.037) $ & $ \mathbf{  0.791 (\pm  0.012) } $ \\
EM & $  0.517 (\pm  0.047) $ & $  0.478 (\pm  0.122) $ & $  0.626 (\pm  0.024) $ \\
Majority Voting & $  0.631 (\pm  0.025) $ & $  0.598 (\pm  0.038) $ & $  0.632 (\pm  0.022) $ \\
\hline
 \multicolumn{4}{|c|}{cybersec} \\
\hline
Simulated consensus & $  0.739 (\pm  0.046) $ & $  0.500 (\pm  0.000) $ & $  0.733 (\pm  0.019) $ \\
Inferred consensus & $  0.500 (\pm  0.000) $ & $  0.500 (\pm  0.000) $ & $  0.699 (\pm  0.071) $ \\
EM & $  0.744 (\pm  0.043) $ & $  0.500 (\pm  0.000) $ & $  0.744 (\pm  0.027) $ \\
Majority Voting & $ \mathbf{  0.773 (\pm  0.015)} $ & $  0.500 (\pm  0.000) $ & $  0.739 (\pm  0.019) $ \\
\hline
 \multicolumn{4}{|c|}{credit-fraud} \\
\hline
Simulated consensus & $  0.704 (\pm  0.117) $ & $  0.500 (\pm  0.000) $ & $  0.737 (\pm  0.052) $ \\
Inferred consensus & $  0.500 (\pm  0.000) $ & $  0.500 (\pm  0.000) $ & $  0.695 (\pm  0.029) $ \\
EM & $  0.732 (\pm  0.130) $ & $  0.500 (\pm  0.000) $ & $  0.688 (\pm  0.011) $ \\
Majority Voting & $  \mathbf{ 0.758 (\pm  0.019)} $ & $  0.500 (\pm  0.000) $ & $  0.687 (\pm  0.012) $ \\
\hline

\end{tabular}
\end{center}
\end{table*}

\section{Active Learning Cycle}

\begin{figure}
    \centering
    \includegraphics[width=4.5in]{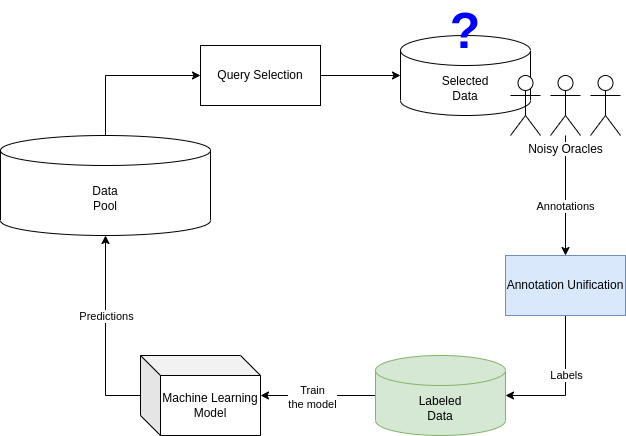}
    \caption{An active learning iterative cycle with noisy annotators. The objectives of this work, i.e. unification algorithm and labeled data, are emphasized with colors.  }
    \label{fig:cycle}
\end{figure}

\end{appendices}